\documentclass[lettersize,journal]{IEEEtran}
\usepackage{amsmath,amsfonts}
\usepackage{algorithmic}
\usepackage{algorithm}
\usepackage{array}
\usepackage[caption=false,font=normalsize,labelfont=sf,textfont=sf]{subfig}
\usepackage{textcomp}
\usepackage{stfloats}
\usepackage{url}
\usepackage{verbatim}
\usepackage{graphicx}
\usepackage{cite}
\usepackage{graphicx}
\usepackage{amsmath}
\usepackage{amssymb}
\usepackage{booktabs}
\usepackage{pgfplots}
\usepackage[mathscr]{eucal}

\usepackage{booktabs}
\usepackage{multirow}
\begin{document}

\title{Towards Unseen Triples: \\Effective Text-Image-joint Learning for Scene Graph Generation}

\author{Qianji Di, \IEEEmembership{}
        Wenxi Ma, \IEEEmembership{}
        Zhongang Qi, \IEEEmembership{}
        Tianxiang Hou, \IEEEmembership{}
        Ying Shan, \IEEEmembership{}
        Hanzi Wang \IEEEmembership{}
      }

\maketitle

\begin{abstract}
Scene Graph Generation (SGG) aims to structurally and comprehensively represent objects and their connections in images, it can significantly benefit scene understanding and other related downstream tasks. Existing SGG models often struggle to solve the long-tailed problem caused by biased datasets. However, even if these models can fit specific datasets better, it may be hard for them to resolve the unseen triples which are not included in the training set. Most methods tend to feed a whole triple and learn the overall features based on statistical machine learning. Such models have difficulty predicting unseen triples because the objects and predicates in the training set are combined differently as novel triples in the test set. In this work, we propose a Text-Image-joint Scene Graph Generation (TISGG) model to resolve the unseen triples and improve the generalisation capability of the SGG models. We propose a Joint Fearture Learning (JFL) module and a Factual Knowledge based Refinement (FKR) module to learn object and predicate categories separately at the feature level and align them with corresponding visual features so that the model is no longer limited to triples matching. Besides, since we observe the long-tailed problem also affects the generalization ability, we design a novel balanced learning strategy, including a Charater Guided Sampling (CGS) and an Informative Re-weighting (IR) module, to provide tailor-made learning methods for each predicate according to their characters. Extensive experiments show that our model achieves state-of-the-art performance. In more detail, TISGG boosts the performances by 11.7\% of zR@20(zero-shot recall) on the PredCls sub-task on the Visual Genome dataset.
\end{abstract}

\section{Introduction}
\begin{figure}[t]
  \centering
  \includegraphics[width=0.9\linewidth]{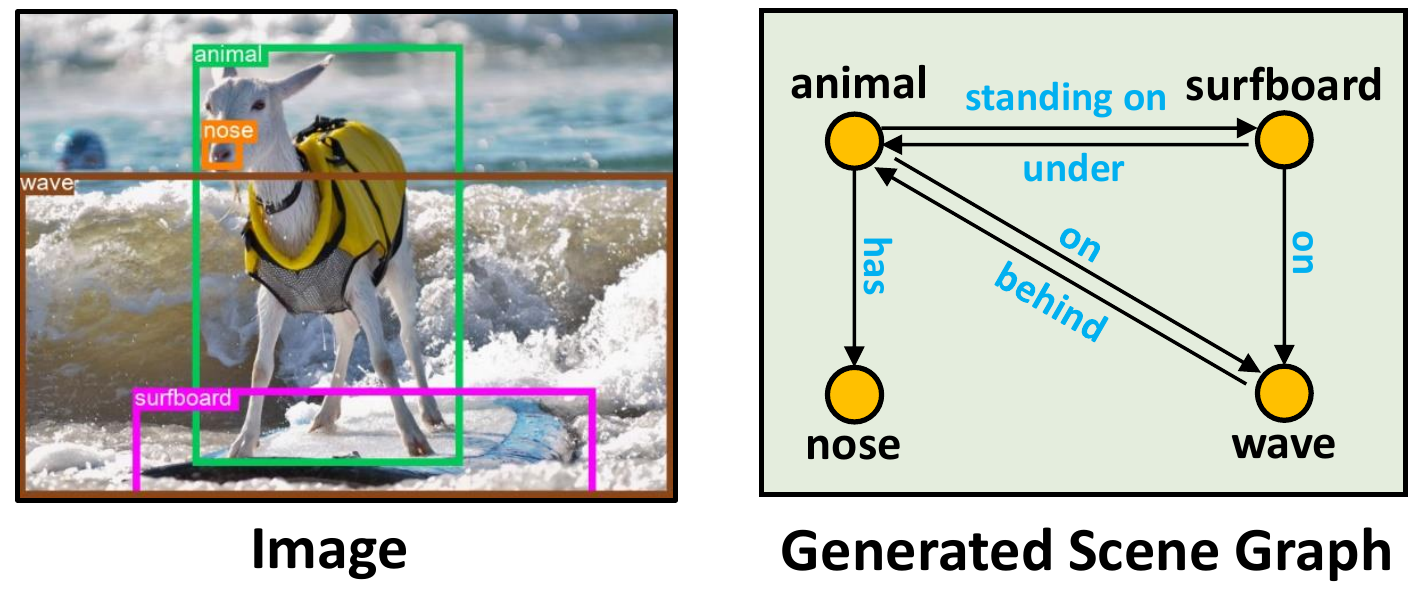}
  \caption{An illustration of an image and its corresponding scene graph. The nodes and edges in the scene graph represent the objects and the relations in the image, respectively.}
\label{fig:intro}
\vspace{-1em}
\end{figure}

\IEEEPARstart{V}{isual} scene understanding and reasoning tasks are usually required to not only detect individual objects but also capture the diverse relations between the objects in images. Simply treating scenes as collections of objects usually fails to predict the semantic relations between objects. Thus, a new data structure called scene graph \cite{johnson2015image} was proposed as a comprehensive and effective representation of scenes. As shown in Figure \ref{fig:intro}, the scene graph describes the original images by representing objects with nodes, relations with edges, resulting in a series of triples (e.g.,  \textit{\textless animal, on, wave\textgreater}).

Similar to knowledge graphs \cite{hogan2021knowledge, joulin2017fast}, such structured representations are more intuitive than representing images with vectors, thus can be easily understood by humans and computers. Scene graphs can be used for many high-level tasks in the computer vision field, such as image retrieval \cite{hildebrandt2020scene,wang2022contrastive}, image captioning \cite{luo2021dual,fei2022attention}, and video question answering (VQA) \cite{teney2017graph,cherian20222,li2022dynamic}, etc. The overall process of scene graph generation (SGG) can be divided into two steps \cite{lu2016visual, DBLP:conf/cvpr/DaiZL17}: detecting objects in given images, and predicting the relations between the detected object pairs. Since the object detection tasks have been well studied, the focus of SGG mainly falls on predicting the relations of the object pairs, i.e., predicates of the triples.

Although SGG techniques have attracted much attention and have evolved significantly since they were proposed, the performance of SGG could be more satisfying for real-world tasks. In addition to the long-tailed problem that existing SGG methods usually care about, SGG suffers from another severe problem of weak generalization ability. Firstly, many existing models usually learn each triple as a whole, or ensemble the triples according to learned knowledge from the predicates, rather than learning the features of each element (including both objects and predicates) in triples individually and ensemble the triples according to these features and factual knowldge. Such models rely heavily on the knowledge learned from the dataset, resulting in poor generalization when predicting triples outside the training set. These triple-level models may ignore the contextual information of the elements within the triples, but try to match the identical triples in the training set. For example, in Figure \ref{fig:intro}, the predicates connecting \emph{animal} and \emph{surfboard} can be naturally proposed by humans with \emph{standing on} based on the the common sense and factual knowledge of the two objects. However, the training set has no triple \textit{{\textless animal, standing on, surfboard\textgreater}}, so existing methods may not make correct predictions. Following this observation, we find that the reason why humans learn better than these models is that we comprehend these object and predicate categories before we try to predict the triple, so we can construct new triples from the feature perspective even if we have not learned the identical combination. We apply this idea to make our model more flexible in predicting new triples. Moreover, we observed that the biased datasets affect the generalization ability, and our feature learning methods can mitigate the long-tailed problem. We additionally introduce factual and contextual knowledge into the prediction, these inherent features will not be influenced by the bias in the datasets, thus our model does not fully dependent on the datasets any more.
For example, the number of triples in the training set of Visual Genome (VG) \cite{krishna2017visual} is about 400,000, of which more than 110,000 contain the predicate $on$, while dozens of other predicates occur only 100 times or less. Due to the long-tailed problem of training samples, the model will be significantly biased to predict high-frequency predicates and is easily influenced by excessive training samples when constructing new triples. However, our feature learning model will consider the factual knowledge from the object and predicate categories instead of the frequencies of the training samples, and propose a contextually related triple. Therefore, the dependency on the datasets is decreased, and the long-tailed effect on the prediction is somewhat mitigated.


To address the above issues, we propose an effective Text-Image alignment Scene Graph Generation (TISGG) model for high-performance scene graph generation.
Firstly, we propose a Joint Feature Learning (JFL) module to align visual features and the corresponding text features to better apply the feature learning. Also, since many words are polysemous, some word embeddings do not accurately point to the meaning that matching the object in the image. This alignment reduces the ambiguity caused by polysemies and improve the performance of our model.
Secondly, we design a Facutal Knowledge based Refinement (FKR) module, which mines the contextual information and factual knowledge of the elements inside triples and refines the predictions. Since these features are inherent, internal, and independent, our model can construct triples by analyzing the object and predicate categories, even if there is no identical combinations in the training set. Furthermore, we propose a Character Guided Sampling (CGS) strategy that resamples the training set guided by the characters of the predicate. We consider the feature and the learning performance of each predicate, and design tailor-made sampling rates for them, achieving a relatively fair predicate distribution. 
In addition, we introduce a model variant TISGG$_{info}$, which employs an Informative Re-weighting (IR) module.  We observe that the classifier is also influenced by the bias, so we utilize Shannon information theory to supervise the weights of the predicates, mitigate the prediction bias, so our model can construct new triples in a more objective manner. 

The main contributions of this work are as follows: 
\begin{itemize}
    \item We identify an essential problem of existing SGG models, they cannot explore the unseen triples though the object and predicate categories are in the training set.
    \item We design a more independent SGG model guided by contextual and factual information for this purpose. Our model learns the internal semantic features and align them with the corresponding visual features to learn the inherent factual knowledge.
    \item We apply a tailor-made sampling strategy for different predicates according to their characters, so that the triple construction process is based on a relatively fair distribution. Furthermore, we propose model variant to prevent the long-tailed problem effecting the triple construction. 
    \item Our model is dataset agnostic with improved generalization ability. Extensive experiments show that our model achieves state-of-the-art performance.
\end{itemize} 

\begin{figure*}[htbp]
  \centering
  \includegraphics[width=\textwidth]{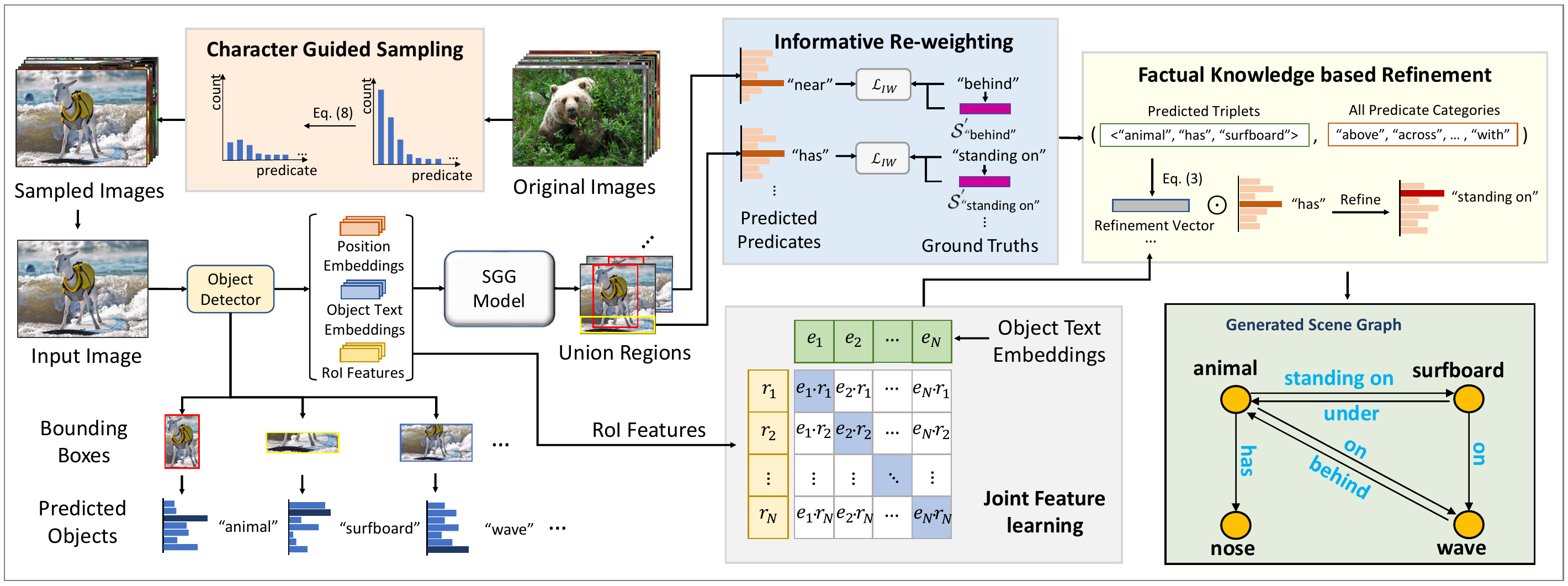}
  \caption{Overview of the proposed TISGG model. Firstly, we resample the dataset with the CGS strategy to get a relatively balanced target domain, from which we get input images and feed them into the pre-trained model for object detection. The object proposals, RoI features, and labels of the images are obtained and fed into our SGG model, where our IR and JFL are implemented. Finally, the obtained triples are passed into the FKR module for refinement to get the final results.}
 \label{fig:pipeline}
\end{figure*}

\section{Related Work}
\noindent
\textbf{Scene Graph Generation.} Scene graphs aim to provide structured representations of images, which were initially proposed for image retrieval \cite{johnson2015image}. Since the Visual Genome (VG) benchmark \cite{krishna2017visual} become publicly available, a large number of methods \cite{lu2016visual, DBLP:conf/cvpr/DaiZL17, zhang2017visual, tang2020unbiased, li2021bipartite, suhail2021energy, guo2021general} have been proposed for scene graph generation (SGG). As research progressed, some methods \cite{xu2017scene, yang2018graph, tang2019learning, chen2019knowledge, zellers2018neural} began to focus on constructing message-passing mechanisms among object pairs and their relations. Recently, some state-of-the-art methods \cite{chen2019counterfactual, tang2020unbiased, yan2020pcpl} aim to alleviate the impact of data bias for SGG. For instance, \cite{tang2020unbiased} utilizes the counterfactual causality to reduce the lousy bias and force the model to focus more on the main visual effects of the relations without losing the context. Unlike existing SGG methods, our proposed TISGG considers the contextual and factual information based on cross-modal feature learning to develop the generalization ability.

\noindent
\textbf{Zero-shot Learning.} Zero-shot learning (ZSL) predicts the classes that have never been seen during training process \cite{xian2018zero}. That is, the classes are not known during supervised learning. Early works on zero-shot learning utilize the features to infer unknown classes. Some recent works \cite{narayan2020latent,li2019leveraging} learn mappings from the image feature space to semantic space to improve ZSL ability. Although some works \cite{zhang2017visual, yang2018shuffle} in computer vision focus on improving ZSL, the metric zero-shot recall was valued very recently in SGG. TDE \cite{tang2020unbiased} introduced zero-shot recall into SGG and found it challenging to be improved based on the VG benchmark. Later on, \cite{suhail2021energy,knyazev2021generative,liu2021fully,goel2022not} discussed the generalization ability of SGG by defining new training losses and data augmentation operations. In this paper, the inherent factual knowledge of labels is utilized to compose triples through the semantic connection between object labels and predicates.

\noindent
\textbf{Multimodal Learning.} The rise of the Internet has facilitated Multimodal Retrieval \cite{rafailidis2013unified,long2016composite}, and machine learning tasks that combine visual and linguistic information have emerged since 2015 \cite{ramachandram2017deep}. Multimodal learning, which aims to process and merge multimodal information, has gradually evolved into an important tool for multimedia analysis and understanding \cite{ochoa2017multimodal}, and researchers have gradually achieved remarkable results in multimodal learning. Common multimodal tasks incorporate mainly speech, text, and visual features. Some language-audio tasks aim to synthesise speech \cite{ning2019review} and captions \cite{drossos2020clotho} from the text. Visual-audio tasks, including visual speech recognition \cite{shillingford2018large} and image/video generation \cite{chen2017deep,wan2019towards} from audio. Several visual-language tasks enable image/video retrieval \cite{elalami2011novel,gabeur2020multi}, image/video captioning \cite{hossain2019comprehensive,wang2018reconstruction}, image/video generation \cite{xu2018attngan,li2018video}, multimodal translation \cite{elliott2017imagination} and human-machine interaction \cite{turk2014multimodal}. There have also been other recent multimodal studies, such as sentiment analysis \cite{soleymani2017survey} and visual grounding \cite{deng2018visual}. In this work, we jointly learn visual and textual features from an image-language perspective to assist scene graph generation.

\section{Approach}
\subsection{Problem Definition}
Formally, given an image $\textbf{I}$, the scene graph $\textbf{S}$ is defined as $\textbf{S}=(\textbf{B},\textbf{O},\textbf{R})$, where $\textbf{B}=\{{b}_i\}_{i=1}^{N_O}$ is the set of bounding boxes, $\textbf{O}=\{o_i\}_{i=1}^{N_O}$ is the set of objects, and $\textbf{R}=\{$\textit \textless$o_i^m,p_i,o_i^n$\textit\textgreater$\}_{i=1}^{M_R}$ is the set of relations between objects. $N_O$ and $M_R$ denote the number of objects and relations in the image, respectively. Specifically, the $i$-th relation in $\textbf{R}$ is represented as \textit\textless$o_i^m,p_i,o_i^n$\textit\textgreater, where $o_i^m$, $o_i^n\in \textbf{O}$ and $p_i\in\textbf{P}$.  $\textbf{P}=\{p_i\}_{i=1}^{N_P}$ is the set of predicates, and $N_P$ denotes the number of predicates. The SGG task $P(\textbf{S}|\textbf{I})$, which aims to improve the performance of predicting $\textbf{S}$ given an input image $\textbf{I}$, is usually decomposed into three components \cite{zellers2018neural}:

\begin{equation}
\label{sgg_qua}
P(\textbf{S}|\textbf{I})=P(\textbf{B}|\textbf{I}) P(\textbf{O}|\textbf{B},\textbf{I}) P(\textbf{R}|\textbf{O},\textbf{B},\textbf{I}),
\end{equation}
where $P(\textbf{B}|\textbf{I})$ and $P(\textbf{O}|\textbf{B},\textbf{I})$ represent the objects localization and prediction, and $P(\textbf{R}|\textbf{O},\textbf{B},\textbf{I})$ denotes the relation prediction between objects. 

\subsection{Overview}
Given an image, current SGG models typically utilize RoI features, bounding boxes, and object labels to predict predicates. Figure \ref{fig:pipeline} shows the overview of our TISGG model. To explore the unseen triples with the seen object and predicate categories in the training set, and improve the generalization ability, our model focuses on learning the linguistic and visual features of the input images.
We first jointly learn the text and image features by aligning the ROI features of the objects and the semantic features of the object and predicate categories. This Joint Feature Learning (JFL) module also reduces the ambiguity caused by polysemy in the object categories.
A Factual Knowledge based Refinement (FKR) module is designed to refine the original predicates. This module considers the inherent factual information of objects and predicate so that the learning process of the model no longer depends only on the training data, thus we make it possible to construct new triple combination. Furthermore, we propose a Character Guided Sampling (CGS) module to mitigate the long-tailed effect in new triple construction process, this is guided by the characters of the predicates. In addition, we also propose a model variant TISGG$_{info}$ which supervises the classifier to support the informative predicates.

\subsection{Joint Feature Learning}
\begin{figure} [t]
	\centering
	\subfloat[\label{fig:4a}]{
		\includegraphics[width=0.22\textwidth]{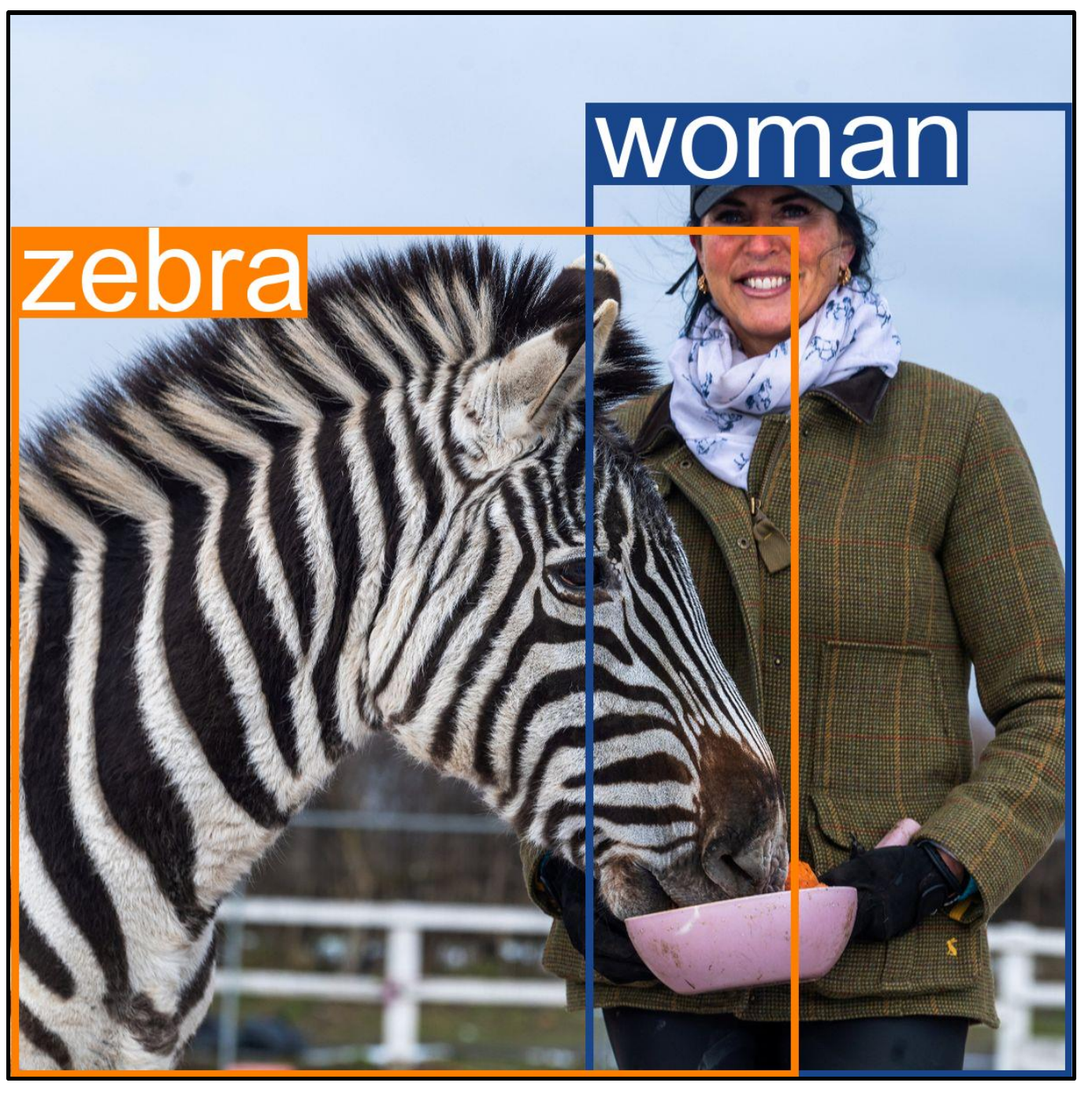}
		}
	\subfloat[\label{fig:4b}]{
		\includegraphics[width=0.22\textwidth]{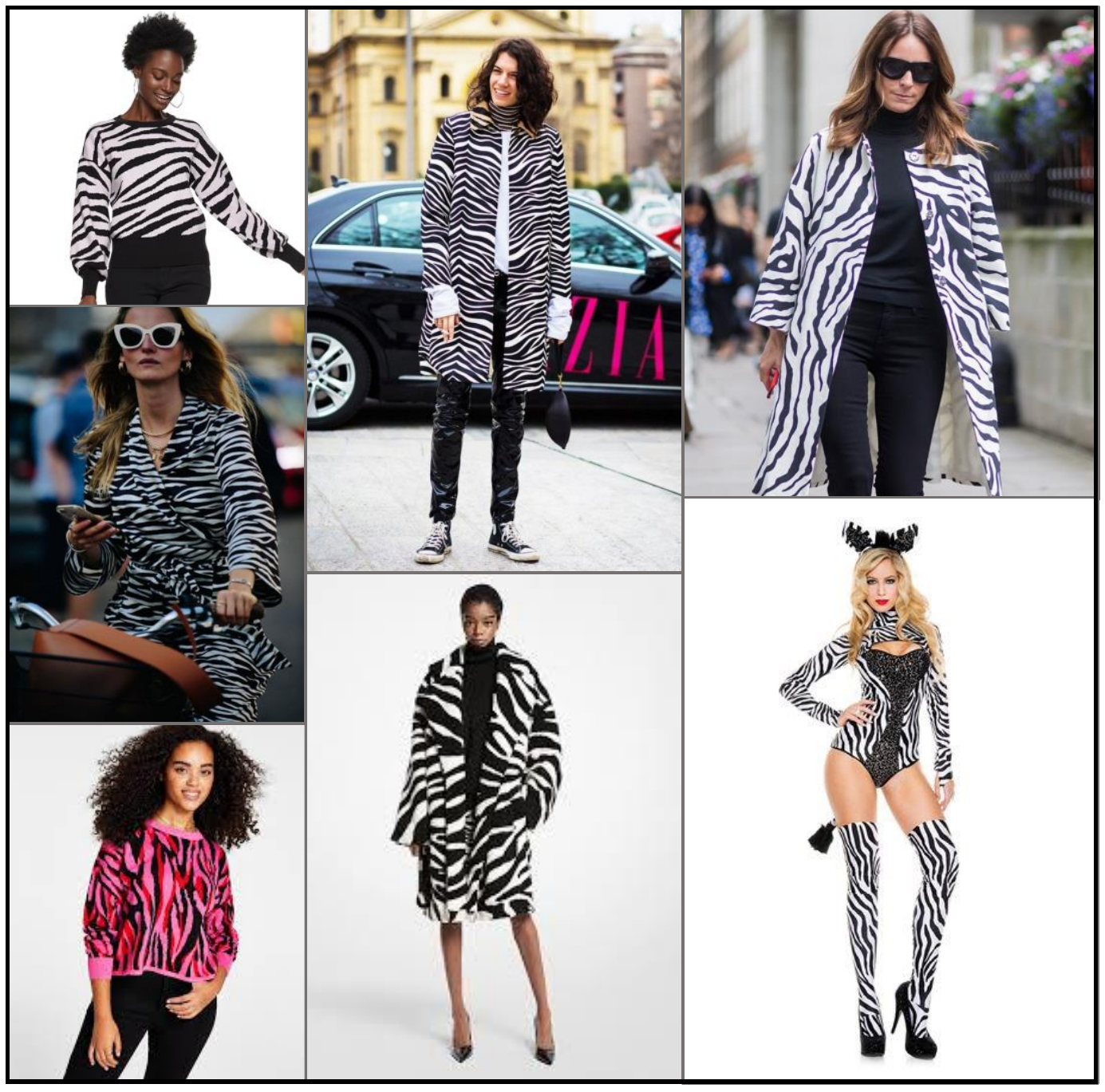}
		}
	\caption{The relation between \emph{woman} and \emph{zebra} is \emph{near} in the original image (a). The image (b) is the output we search the Internet with the keywords \emph{zebra} and \emph{woman}.}
		\label{fig:align}
\end{figure}

In order to construct unseen triples, we will learn the features from each element in the known samples. The unseen combinations and the learned triples will be treated alike during the prediction process in our model, we try to construct every triples regardless of the frequency of it in the training set, so unseen combinations are also possible to be predicted. We first learn the elements inside the triples separately, them ensemble the knowledge to predict the final triples. Our model learns the individual features of the elements in triples, including the object and predicate categories features (text features), and the corresponding visual features. The text features can introduce the factual knowledge, and the visual features can significantly help the visual understanding. During the learning process, we extract and align these image and text features, gather the joint knowledge to predict the combination.
What's more, when we conduct experiments on FKR (which will be described in the next subsection), we found a problem by analyzing the results. FKR utilizes semantic information of labels to assist the inference procedure, but semantic information can sometimes introduce ambiguity. Specifically, many words are polysemous, but an image should have a single intention only. The FKR module cannot guarantee a correct matching between the word embedding and the objects in the image when the label is polysemous. We notice some unusual results in the output. For example, for the two objects in Figure \ref{fig:4a}, the highest ranking in the results given by FKR turned out to be \textit{\textless woman, wearing, zebra \textgreater} and \textit{\textless woman, in, zebra \textgreater}. These triples do not fit common sense understanding or frequency-based prediction. We analyzed that these semantic-based word embeddings are generated considering polysemy. Apart from the meaning "zebra the animal", the word \emph{zebra} also means "stripes" and "crosswalk". We searched for the words \emph{woman} and \emph{zebra} online (Figure \ref{fig:4b}), and the results also show that "a woman wearing striped clothing" is a natural association for these two words, so FKR would prefer \emph{wearing} and \emph{in} as the results. However, SGG works with given images, such ambiguity can lead to many mistakes. We need to align image features with semantic features to reduce ambiguity.

We consider using contrastive learning. It is worth noting that, unlike previous CV models with contrastive learning, we focus on the visual features of the regions of objects, i.e., the RoI features, because we need to find the specific visual features corresponding to the object categories instead of processing the whole image. The word embedding of the object label and corresponding object visual feature (RoI feature of the object proposal) in the image are used as a positive pair, and the rest are negative pairs. The model learns to increase the similarity between the positive pairs while decreasing the similarity between the negative pairs by contrastive loss. We use $r_i$ and $e_i$ to denote the RoI feature and the semantic word embedding of $o_i\in \textbf{O}$ and leverage cosine similarity to measure the similarity between RoI features and semantic features:
\begin{equation}
sim(r_i,e_j)  =  \frac{e_jr_i}{\Vert e_j\Vert \Vert r_i\Vert},
\end{equation}
and the contrastive loss is then formulated by:
\begin{align}
& \mathscr{L}_{r2e} = -\frac{1}{N_O}\sum_i^{N_O}{log\frac{\text{exp}(sim(r_i,e_i))}{\sum_j^N \text{exp}(sim(r_i,e_j))}}, \\
& \mathscr{L}_{e2r} = -\frac{1}{N_O}\sum_i^{N_O}{log\frac{\text{exp}(sim(r_i,e_i))}{\sum_j^N \text{exp}(sim(r_j,e_i))}}, \\
& \mathscr{L}_c = \frac{1}{2} (\mathscr{L}_{r2e} + \mathscr{L}_{e2r}),
\end{align}
where $\mathscr{L}_{r2e}$, $\mathscr{L}_{e2r}$ denote image-to-text and text-to-image loss, and $\mathscr{L}_{c}$ is the final contrastive loss. We add the contrastive loss and the prediction loss from the FKR module as the total loss. The model will adjust the prediction of the predicates according to the total loss based on the feature-level learning of the elements inside triples. At this point, we get the image features aligned with the labels, and the FKR module will further learn the predicate prediction based on the adjusted features. We train image-text alignment in a self-supervised way using contrastive loss.


\subsection{Factual Knowledge based Refinement}
To improve the generalization ability, we design a Factual Knowledge based Refinement (FKR) module, which considers the intrinsic semantic information and the factual knowledge of object and predicate categories. These features are neither determined nor influenced by any particular dataset. On the one hand, referring to such information during the learning process reduces the dependence on datasets, the model will be less adversely affected. On the other hand, learning the intrinsic features of categories is useful for constructing triples. Our FKR module employs these features and gather the factual knowledge of the categories to assist our model. By this means, our model's dependence on the dataset is reduced, and the generalization ability is improved.

\begin{figure}[t]
  \centering
  \includegraphics[width=0.45\textwidth]{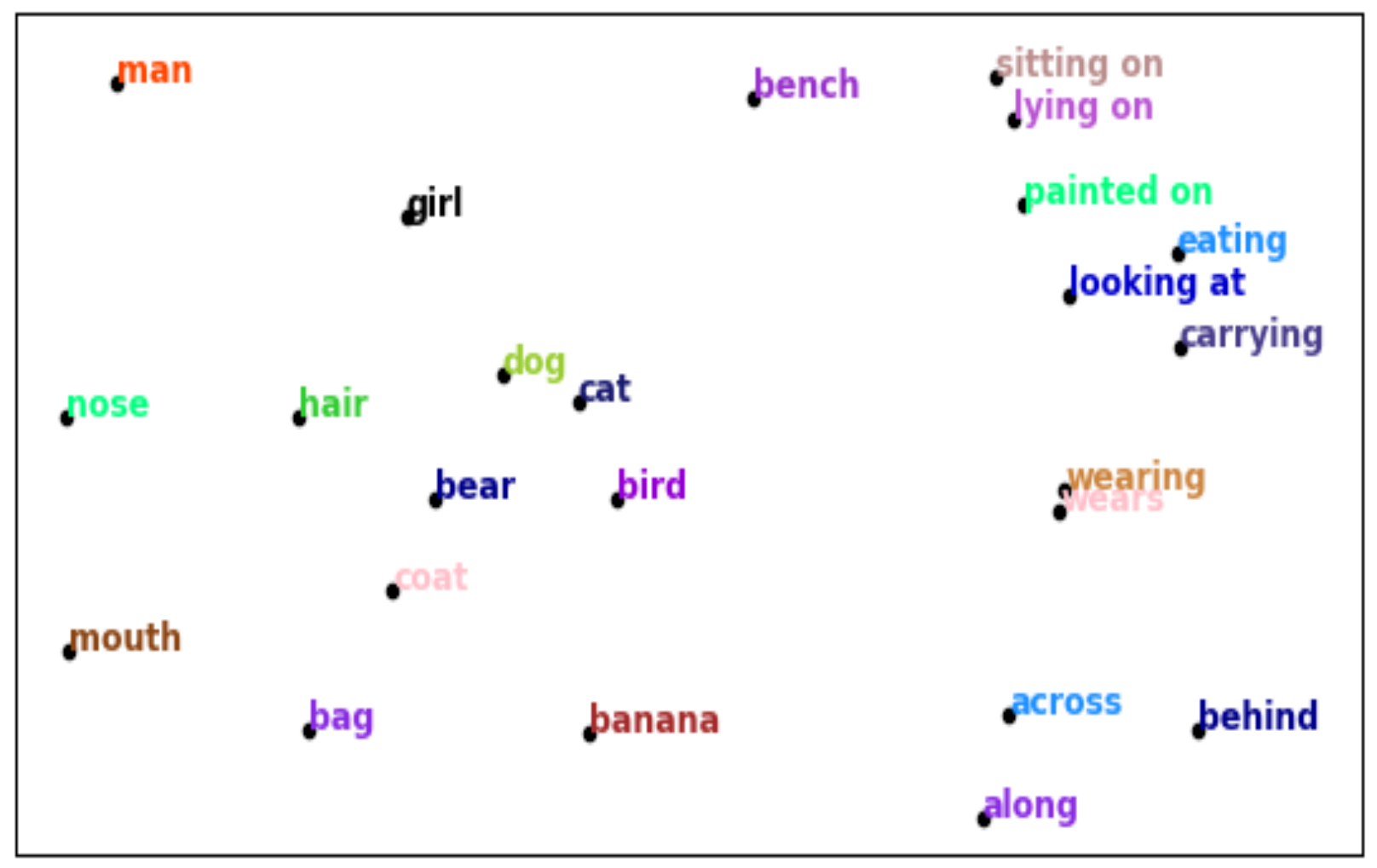}
  \caption{The t-SNE visualization of the word embeddings. For clarity, we only list a few typical word embeddings here.}
\label{fig:sd}
\end{figure}

Our TISGG introduces factual information using the word embeddings learned by \emph{Bert} \cite{devlin2018bert}. \emph{Bert} is currently one of the most popular word embedding models, it generates word embedding distribution according to the semantics of words. In Figure \ref{fig:sd}, we visualize the \emph{Bert} distribution by the t-SNE \cite{van2008visualizing} technique. It can be seen that the vectors of words with higher semantic similarity are closer to each other in the vector space (e.g., \emph{wearing} and \emph{wears}) and vice versa. For example, in many previous SGG models, it is challenging to form triples with \emph{bench} and \emph{sitting on} or \emph{lying on} due to the limitation of quantities of the corresponding training samples (8003 and 561 out of 400,000). \emph{Bench} is always coupled with \emph{on} in the training set, in which case it is difficult to generate phrases like \emph{sitting on bench} in the prediction stage. However, after we introduce factual knowledge, the connection between objects and predicates like \emph{bench} and \emph{sitting on} becomes more robust because they are more related in commonsense, and are very close in the vector space. For the model, we want the probability of matching \emph{sitting on} to the object \emph{bench} grows, and the probability of \emph{on} to decrease. The proposed FKR refines the originally predicted predicates by utilizing the factual knowledge of predicate and object categories to obtain more reasonable results. 

We denote $\mathcal{P}$ as the predicate set, which contains all the categories of predicates in the dataset. 
Then, given the original relation prediction $\textless \hat{o}_i^m,\hat{p}_i,\hat{o}_i^n\textgreater$ and the predicate distribution $\mathcal{D}_i\in \mathbb{R}^C$, we utilize a refinement vector $\textbf{v}_i\in \mathbb{R}^C$ to generate the final refined predicates $\bar{p}_i$:
\begin{equation}
\bar{p}_i = \mathcal{F}(\mathop{\arg\max}\limits_{i}(\mathcal{D}_i \odot\textbf{v}_i)),
\end{equation}
\begin{equation}
\label{alpha}
\textbf{v}_i=\alpha (d(\hat{o}_i^m,\mathcal{P})+d(\hat{o}_i^n,\mathcal{P})) + (1-\alpha) d(\hat{p}_i,\mathcal{P}),
\end{equation}
Where $\odot$ denotes the Hadamard product and $\mathcal{F}(\cdot)$ is a mapping function that maps the predicate index to its corresponding predicate. $d(a,B)$ denotes the Euclid distance of the vector $a$ and all the vectors in the set $B$.

We take an original predicted triple \textit{\textless girl,\ painted\ on,\ bench\textgreater} and its ground truth \textit{\textless girl,\ sitting\ on,\ bench\textgreater} as an example. FKR will consider the semantic distance between objects and predicates, i.e., $d(girl,\ sitting\ on)+d(bench,\ sitting\ on)$ is smaller than $d(girl,\ painted\ on)+d(bench,\ painted\ on)$, so \emph{painted on} will be refined to \emph{sitting on}. At the same time, FKR will also consider the semantic distance between the predicates, i.e., when FKR refines a predicate, those predicates with similar semantics will also be predicted with high probability. For example, in Figure \ref{fig:sd}, \emph{sitting\ on} is also possible to be refined to \emph{lying\ on}. These newly constructed triples tend to meet the commonsense, and are free from the restriction of the dataset.

%

\subsection{Balanced Learning Strategy}
Since the long-tailed problem significantly impacts the model's generalization ability, we further explore some balanced learning strategies.

\begin{figure}[htbp]
  \centering
  \includegraphics[width=0.48\textwidth]{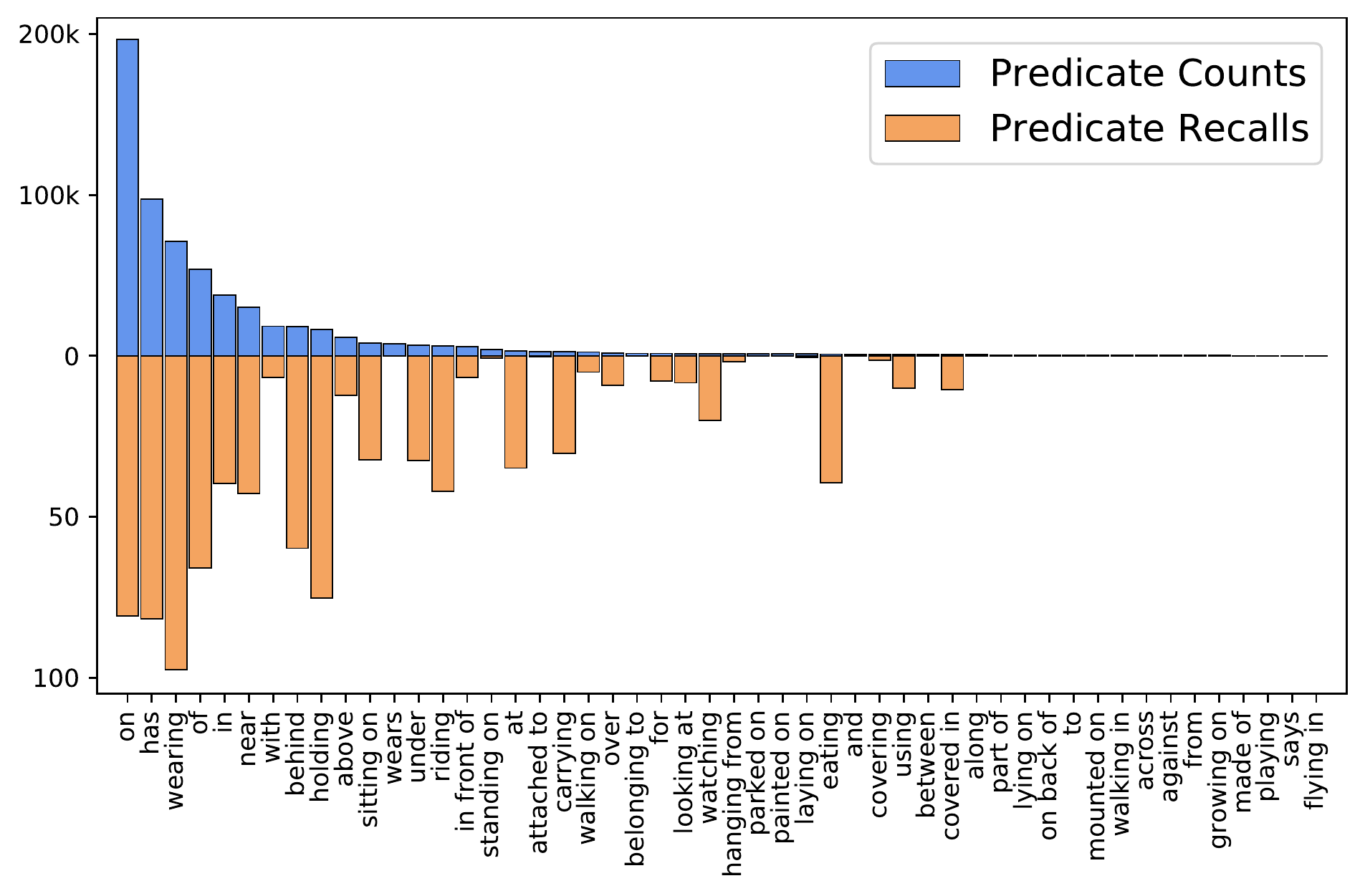}
  \caption{A visualization of the baseline recalls and frequency counts of predicates. The blue bars represent the counts of the predicates in the training set. The orange bars represent the recall of each predicate in the baseline model.}
\label{fig:fig5}
\end{figure}

\noindent
\textbf{Character Guided Sampling.} An intuitive approach is to down-sample the head predicates to reduce the impact, and make the dataset more balanced. However, after conducting the baseline model, we noticed some head predicates with low recalls, which means that even though these predicates are numerous, the model needs help understanding their semantic and factual knowledge well. For example, both \emph{with} and \emph{attached to} are head predicates, but their recalls are much lower than other head predicates (See Figure \ref{fig:fig5}). These predicates are more abstract, even if there are plenty of training samples, the model still cannot manage these predicates and prefers to choose other head predicates that may fit the situation. In this case, if we cut down the number of all head predicates, the situation will be even worse for such complex predicates, and the distribution is not fair enough. As shown in Figure \ref{fig:fig5}, the baseline recalls of predicates are not positively correlated with the counts. Therefore, when we consider cutting down the numbers of the head predicates, we should take care of those with low recalls simultaneously. Based on this observation, we present the Character Guided Sampling (CGS) strategy to balance the impact of the datasets. Specifically, we sample $n_i$ triples for the predicate of the $i_{th}$ category from the original training set. $n_i$ is computed by $n_i = \mathcal{N}_i\times s_i$, where $\mathcal{N}_i$ and $s_i$ denote the number and the sampling rate of the $i$-th predicate, respectively. Formally, the sampling rate $s_i$ is defined as:
\begin{equation}
\label{gs}
s_i = \begin{cases}
\min\nolimits{(\frac{\tau}{\mathcal{N}_i\ \times \ \beta c_i},1)}, & \mathcal{N}_i \ge \tau \\
1,& others \\
\end{cases},
\end{equation}
where $c_i$ is the predicate recall of the $i$-th category,  $\tau$ is the threshold of head predicates, and $\beta$ is a scalar factor. By this means, $s_i$ is jointly determined by $\mathcal{N}_i$ and $c_i$. CGS is a model-independent sampling strategy and is not designed for any particular dataset so that it can be widely used.

In our CGS, we down-sample those head predicates with high recalls and retain other head predicates with low recalls. Notably, up-sampling is not considered in our work because duplicate samples bring little information to the model and may increase the risk of over-fitting.

\noindent
\textbf{Informative Re-weighting.} General SGG process \cite{tang2019learning,tang2020unbiased} are usually trained under the supervision of three losses: the object localization loss $\mathcal{L}_{box}$, the object classification loss $\mathcal{L}_{obj}$, and the predicate classification loss $\mathcal{L}_{pred}$. The final loss is computed as:
\begin{equation}
\label{step1}
\mathcal{L}= \mathcal{L}_{box}+\mathcal{L}_{obj}+\mathcal{L}_{pred}
\end{equation}
However, some general but not informative predicates account for a large part of the training set, so it is not appropriate to learn the different predicates equally. To penalize predicates differently, we assign weights to $\mathcal{L}_{pred}$ during back-propagation. Specifically, we calculate the Shannon information content $\mathcal{S}_{p_i}$ for predicate $p_i$. Then we take the normalized Shannon information content $\mathcal{S}_{p_i}^{'}$ as the weight of the predicate classification loss $\mathcal{L}_{pred}$ for $p_i$. The weighted loss is denoted by $\mathcal{L}_{IW}$, which can be defined as follows:
\begin{equation}
\label{iw}
\mathcal{L}_{IW}=-\sum\nolimits_{i=1}^{M}\mathcal{S}_{p_i}^{'}\hat p_ilogp_i.
\end{equation}

We replace the original predicate classification loss $\mathcal{L}_{pred}$ with $\mathcal{L}_{IW}$ and apply a weight $\mu$ to it for the training. Finally, the total loss $\mathcal{L}_{TISGG}$ can be defined as follows:
\begin{equation}
\label{step2}
\mathcal{L}_{TISGG}= \mathcal{L}_{box}+\mathcal{L}_{obj}+\mathcal{L}_{c}+\mu \mathcal{L}_{IW}.
\end{equation}

By utilizing the information content of predicates, the proposed TISGG$_{info}$ is promoted to pay more attention to the informative predicates. The proposed $\mathcal{L}_{IW}$ can significantly alleviate the negative impact brought by the long-tailed training set.

\section{Experiments}

\subsection{Experiment Settings}
Following the general approaches \cite{tang2019learning,xu2017scene, yang2018graph, chen2019knowledge, zellers2018neural,chen2019counterfactual, tang2020unbiased}, the evaluation is conducted on three subtasks of SGG: Predicate Classification (PredCls), Scene Graph Classification (SGCls), and Scene Graph Generation (SGGen). The three settings are sequentially getting more difficult.

\begin{table*}[]
\caption{the zR@K comparison of existing methods and our TISGG on the PredCls, SGCls, and SGGen subtasks. The best result in each column is \textbf{bolded}. }
\label{tab:quantitative}
\scalebox{1.19}{
\begin{tabular}{c|ccc|ccc|ccc}
\hline
\multirow{2}{*}{Methods}                                 & \multicolumn{3}{c|}{PredCls}                  & \multicolumn{3}{c|}{SGCls}                 & \multicolumn{3}{c}{SGCls}                  \\ \cline{2-10} 
                                                         & zR@20         & zR@50         & zR@100        & zR@20        & zR@50        & zR@100       & zR@20        & zR@50        & zR@100       \\ \hline
TDE \cite{tang2020unbiased}              & -             & 5.9           & 8.1           & -            & 3.0          & 3.7          & -            & 2.2          & 2.8          \\
VCTree   \cite{tang2019learning}         & 2.2           & 5.4           & 10.2          & 0.8          & 2.6          & 2.6          & 0.0         & 0.3          & 0.8          \\
Motif \cite{zellers2018neural}           & 1.2           & 3.5           & 10.3          & 0.3          & 0.6          & 1.1          & 0.0          & 0.0          & 0.1          \\
Transformer  \cite{vaswani2017attention} & 1.7           & 6.7           & 10.6          & 1.8          & 2.6          & 3.9          & 0.0          & 0.0          & 0.0          \\
EBM (VCTree) \cite{suhail2021energy}     & 2.2           & 5.3           & -             & 0.8          & 1.8          & -            & 0.2          & 0.5          & -            \\
BASGG (VCTree)  \cite{guo2021general}    & 2.9           & 5.5           & 7.5           & 1.5          & 3.5          & 5.0          & 0.0          & 0.0          & 0.9          \\
Equal (VCTree)  \cite{goel2022not}       & 1.5           & 3.7           & -             & 0.3          & 1.0          & -            & 0.4          & 0.9          & -            \\
FCSGG  \cite{liu2021fully}               & -             & 8.2           & 10.6          & -            & 1.3          & 1.7          & -            & 0.8          & 1.1          \\ \hline
TISGG                                                    & \textbf{13.6} & \textbf{20.2} & \textbf{22.3} & \textbf{5.8} & \textbf{7.4} & \textbf{8.4} & \textbf{2.9} & \textbf{4.4} & \textbf{5.1} \\
 \hline
\end{tabular}}
\end{table*}

\noindent
\textbf{Dataset.}
Following the previous works \cite{zellers2018neural,tang2019learning, tang2020unbiased}, we mainly evaluate our TISGG on the Visual Genome (VG) dataset \cite{krishna2017visual}. The original VG dataset contains 108k images, including 75k object categories and 37k predicate categories. 
Since VG is pretty sparse, we adopt a split following the previous works \cite{xu2017scene,tang2019learning,yang2018graph,tang2020unbiased,guo2021general} called VG-150, a subset of the original VG dataset that contains the 150 most frequent object categories, 50 predicate categories, and the same number of images. In the experiments, the dataset has been divided into the training set (70\%) and the test set (30\%). In addition, we follow \cite{zellers2018neural,tang2020unbiased,guo2021general} to split a validation set (5k) from the training set and a zero-shot test set (5k). The zero-shot test set consists of novel combinations with objects and predicates from the training set. In order to verify the generalization ability of our TISGG, we also conduct training on the Open Images V4 and V6 datasets following previous works \cite{li2021bipartite, zhang2019graphical,lin2020gps, kuznetsova2020open}. Since these two datasets have no zero-shot test sets, we compare the other metrics of the model on these two datasets.

\noindent
\textbf{Metrics.}
In order to show the ability of exploring unseen triples of the proposed TISGG, we report the zero-shot Recall (zR@K) \cite{lu2016visual} in experiments. zR@K computes R@Ks for the triples that have never been seen in the training set. Besides, we evaluated Recall@K (R@K) and mean Recall@K (mR@K) to validate the overall performance of the model. R@K averages the recalls for all samples, while mR@K averages the recalls across all predicate categories. Since R@K ignores the contributions of different predicates, we mainly consider mR@K to validate the effectiveness of our methods on the long-tailed problem. In addition, we adopt a new metric, mRIC@K \cite{guo2021general}. This metric is calculated by multiplying the recall with information content of each predicate, reflecting how much information contained in the generated scene graphs.

\begin{table*}[]
\caption{Balanced learning results comparison of our TISGG and other exsiting methods. The best result of each column is \textbf{bolded}.}
\label{tab:mr}
\scalebox{1.08}{
\begin{tabular}{c|ccc|ccc|ccc}
\hline
\multirow{2}{*}{Methods}                                                        & \multicolumn{3}{c|}{PredCls} & \multicolumn{3}{c|}{SGCls} & \multicolumn{3}{c}{SGGen} \\ \cline{2-10} 
                                                                                & mR@20   & mR@50   & mR@100   & mR@20   & mR@50  & mR@100  & mR@20  & mR@50  & mR@100  \\ \hline
 TDE  \cite{tang2020unbiased}                                & 18.5    & 25.5    & 29.1     & 11.1    & 13.9   & 15.2    & 6.6    & 8.5    & 9.9     \\
 FCSGG  \cite{liu2021fully}                                  & 4.9     & 6.3     & 7.1      & 2.9     & 3.7    & 4.1     & 2.7    & 3.6    & 4.2     \\
 PUM  \cite{yang2021probabilistic}                           & -       & 20.2    & 22.0     & -       & 11.9   & 12.8    & -      & 7.7    & 8.9     \\
 BGNN  \cite{li2021bipartite}                                & -       & 30.4    & 32.9     & -       & 14.3   & 16.5    & -      & 10.7   & 12.6    \\
 PPDL  \cite{li2022ppdl}                                     & -       & 33.0    & 36.2     & -       & 20.2   & 22.0    & -      & 12.2   & 14.4    \\
 MotifNet  \cite{zellers2018neural}                          & 11.5    & 14.6    & 15.8     & 6.5     & 8.0    & 8.5     & 4.1    & 5.5    & 6.8     \\
 BA-SGG (MotifNet)  \cite{guo2021general}                    & 24.8    & 29.7    & 31.7     & 14.0    & 16.5   & 17.5    & 10.7   & 13.5   & 15.6    \\
 VCTree  \cite{tang2019learning}                             & 11.7    & 14.9    & 16.1     & 6.2     & 7.5    & 7.9     & 4.2    & 5.7    & 6.9     \\
 BA-SGG (VCTree)  \cite{guo2021general} & 26.2    & 30.6    & 32.6     & 13.9    & 16.2   & 17.1    & 10.5   & 13.5   & 15.7    \\
Transformer  \cite {vaswani2017attention}                        & 12.4    & 16.0    & 17.5     & 7.5     & 9.2    & 9.9     & 5.1    & 7.1    & 8.5     \\
 BA-SGG (Transformer)  \cite{guo2021general}                 & 22.4    & 27.4    & 29.4     & 15.7    & 18.5   & 19.4    & 11.4   & 14.8   & 17.1    \\ \hline
TISGG                                                                           & 29.2    & 35.1    & 37.5     & 15.9    & 19.2   & 20.2    & 11.7   & 15.4   & 17.9    \\
TISGG$_{info}$                                                                  & \textbf{30.1}    & \textbf{36.1}    & \textbf{38.4}     & \textbf{17.7}    & \textbf{20.4}   & \textbf{21.5}    & \textbf{13.0}   & \textbf{16.6}   & \textbf{18.9}    \\ \hline
\end{tabular}}
\end{table*}

\noindent
\textbf{Implementation Details.}
Following the experimental settings of the previous works \cite{tang2020unbiased,guo2021general}, we adopt Faster R-CNN \cite{ren2015faster} as our object detector, and ResNeXt-101-FPN \cite{lin2017feature} as the backbone network. The parameters of the object detector are fixed during the training phase. We train the whole network with the maximum iterations of 50k. The initial learning rate is set to 0.001, and it decays by a factor of 10 when the validation performance plateaus. The $\alpha$ in FKR is a scalar parameter, which is empirically set to 0.35 in our experiments. The threshold $\tau$ and the scalar factor $\beta$ are set as 1100 and 0.3, respectively. $\mu$ is set to 1.2 in our experiments.

\begin{table}[]
\normalsize
\caption{PredCls subtask R@K and mRIC@K results comparison of some existing methods and our TISGG.}
\label{tab:overall}
\scalebox{0.8}{
\begin{tabular}{c|cc|cc}
\hline
Methods                                                                         & R@50 & R@100 & mRIC@50 & mRIC@100 \\ \hline
 MotifNet  \cite{zellers2018neural}                          & 65.7 & 67.7  & 50.6    & 56.9     \\
 BA-SGG (MotifNet)  \cite{guo2021general}                    & 36.4 & 38.1  & 158.7   & 171.2    \\
 TISGG (MotifNet)                                                             & 46.0 & 47.3  & 175.3   & 184.8    \\ 
 TISGG$_{info}$ (MotifNet)                                                             & 44.6 & 45.8  & 195.8   & 194.5    \\\hline
VCTree  \cite{tang2019learning}                             & 66.0 & 67.8  & 57.4    & 64.1     \\
 BA-SGG (VCTree)  \cite{guo2021general} & 47.3 & 49.1  & 156.8   & 167.2    \\
 TISGG (VCTree)                                                               & 49.5 & 50.6  & 175.1   & 183.9    \\
 TISGG$_{info}$ (VCTree)                                                               & 47.6 & 47.8  & 193.4   & 199.2    \\\hline
Transformer  \cite{vaswani2017attention}                        & 65.5 & 67.4  & 53.5    & 59.2     \\
 BA-SGG (Transformer)  \cite{guo2021general}                 & 46.9 & 48.6  & 134.8   & 142.8    \\
 TISGG (Transformer)                                                         & 47.5 & 49.1  & 185.9   & 198.8    \\
 TISGG$_{info}$ (Transformer)                                                         & 47.2 & 48.8  & 195.5   & 211.8    \\\hline
\end{tabular}}
\end{table}

\subsection{Quantitative Results}
\noindent
\textbf{Unseen Triples. }Following \cite{tang2020unbiased}, we use zR@K to verify if our TISGG could construct unseen triples. We list some state-of-the-art SGG methods that also display zR@K to compare the results. As seen from Table \ref{tab:quantitative}, exploring novel triples is quite challenging to existing SGG methods. 
Compared with some existing works, TISGG has a more significant improvement in zR@K, this is because our model lays stress on feature learning from the perspective of the text and image. For example, TISGG is about 10 times more accurate in PredCls than other models. Even if SGGen task is challenging with uncertain object proposals and labels, our model can still predict more correct triples compared to other methods. Our TISGG is capable of constructing novel triples based on the comprehensive factual knowledge learned from the individual elements in the training set. What's more, some recent works with balanced learning strategies may have slightly lower zR@K, they need to be more adaptable to the unbalanced datasets to overcome the bias. However, fitting one specific dataset may lead to poor generalization ability. Our TISGG is independent of datasets to some extent because we utilize the intrinsic features, so we can predict both seen and novel triples. 

\noindent
\textbf{Overall performance. } Since our TISGG learns from a feature perspective, our zR@K results are much improved, while other metrics such as mR@K and R@K are still competitive. Our mR@K results are shown in Table \ref{tab:mr}. We learn to construct triples independently, so the long-tailed effect of VG is mitigated a lot. Our TISGG outperforms many existing methods in mR@K, and TISGG$_{info}$ even improves it more since EGL supervises the classifier to generate more informative results. The overall performance of our model is shown in Table \ref{tab:overall}. It is worth noting that our R@K is not outperforming some of the existing methods. This is because our model improves the prediction probability of many low-frequency predicates, thus the probabilities of head predicates such as \emph{on} decrease. Since head predicates significantly impact R@K, it will be easily improved when the model chooses more head predicates, so R@K is not the best metric in VG benchmark. What's more, our mRIC@K is almost three times as much as the models with high R@K, indicating that our model is not only more balanced, but can also predicate more informative and practical results. Compared with some balanced learning methods (BASGG \cite{guo2021general}), our model gets both higher R@K and mRIC@K.

\noindent
\textbf{Generalization. }To verify the generalization ability of our model, we also conduct experiments on different datasets. As shown in Table \ref{tab:v4}, our model can produce competitive results on both Open Images V4 and V6 datasets. Our R@K outperforms the existing models, indicating that our model works better on other datasets than these methods. We additionally compare the weighted mean AP of relationship (rel) and phrase (phr) according to the Open Images settings following \cite{li2021bipartite}. It can be seen that our TISGG, although aiming to improve generalization ability rather than targeting improving recalls, is still competitive with these methods. 

\begin{table}[]
\caption{Experiments on Open Images datasets on SGGen subtask.}
\label{tab:v4}
\scalebox{1.13}{
\begin{tabular}{c|c|cc|cc}
\hline
\multirow{2}{*}{Datasets} & \multirow{2}{*}{Methods} & \multirow{2}{*}{R@50} & \multirow{2}{*}{R@100} & \multicolumn{2}{c}{wmAP} \\ \cline{5-6} 
                          &                          &                       &                        & rel         & phr        \\ \hline
\multirow{4}{*}{V4}       & RelDN\cite{zhang2019graphical}                    & 75.6                  & -                      & 35.5        & 38.5       \\
                          & BGNN \cite{li2021bipartite}                    & 75.5                  & 77.2                   & 37.7        & 41.7       \\ \cline{2-6} 
                          & TISGG                    & 75.0                  & 76.5                   & 35.3        & 38.9       \\
                          & TISGG$_{info}$                & 75.9                  & 78.6                   & 35.8        & 39.8       \\ \hline
\multirow{5}{*}{V6}       & RelDN \cite{zhang2019graphical}                   & 73.0                  & -                      & 32.1        & 33.4       \\
                          & VCTree \cite{tang2019learning}                  & 74.0                  & 75.1                   & 34.2        & 33.1       \\
                          & BGNN \cite{li2021bipartite}                    & 74.9                  & 76.5                   & 33.5        & 34.2       \\ \cline{2-6} 
                          & TISGG                    & 75.2                  & 75.7                   & 31.2        & 33.5       \\
                          & TISGG$_{info}$                & 76.0                  & 78.2                   & 34.0        & 34.8       \\ \hline
\end{tabular}}
\end{table}

\begin{table}[t]
  \caption{Ablation experimental results of our modules on the Visual Genome dataset.}
  \label{tab:ablation}
  \scalebox{0.9}{\tabcolsep 2mm
  \begin{tabular}{c|ccc|ccc}
    \toprule
    {Methods} & {zR@20} & {zR@50} & {zR@100} & {mR@20} & {mR@50} & {mR@100}\\ 
    \hline
    {VCTree\cite{tang2019learning}} & 2.2 & 5.4 & 10.2 & 11.7 & 14.9 & 16.1 \\
    {JFL} & 5.9 & 9.7 & 10.6 & 14.7 & 17.6 & 18.8 \\
    {FKR} & 7.9 & 12.1 & 14.4 & 15.9 & 18.9 & 20.3 \\
    {CGS} & 5.3 & 8.0 & 9.9 & 18.9 & 22.7 & 24.3 \\
    {IR} & 3.5 & 6.8 & 9.0 & 18.7 & 23.1 & 25.9 \\ \hline
    
    {FKR + JFL} & 11.4 & 18.2 & 19.2 & 18.7 & 21.6 & 23.7 \\
    {FKR + CGS} & 8.7 & 13.1 & 15.4 & 21.1  & 26.2  & 27.7 \\
    {JFL + CGS} & 10.5 & 16.4 & 17.2 & 22.6 & 28.1 & 30.2 \\
    {IR + CGS} & 6.4 & 11.6 & 14.2 & 27.1 & 32.1 & 35.2 \\
    {FKR + IR} & 5.8 & 8.7 & 9.8 & 20.2 & 27.2 & 27.6 \\
    {JFL + IR} & 8.8 & 12.3 & 12.6 & 21.3 & 26.8 & 29.9 \\ 
    \bottomrule
  \end{tabular}}
\end{table}

\begin{figure*} [t]
	\centering
	\subfloat[\label{fig:7a}]{
		\includegraphics[width=0.5\textwidth]{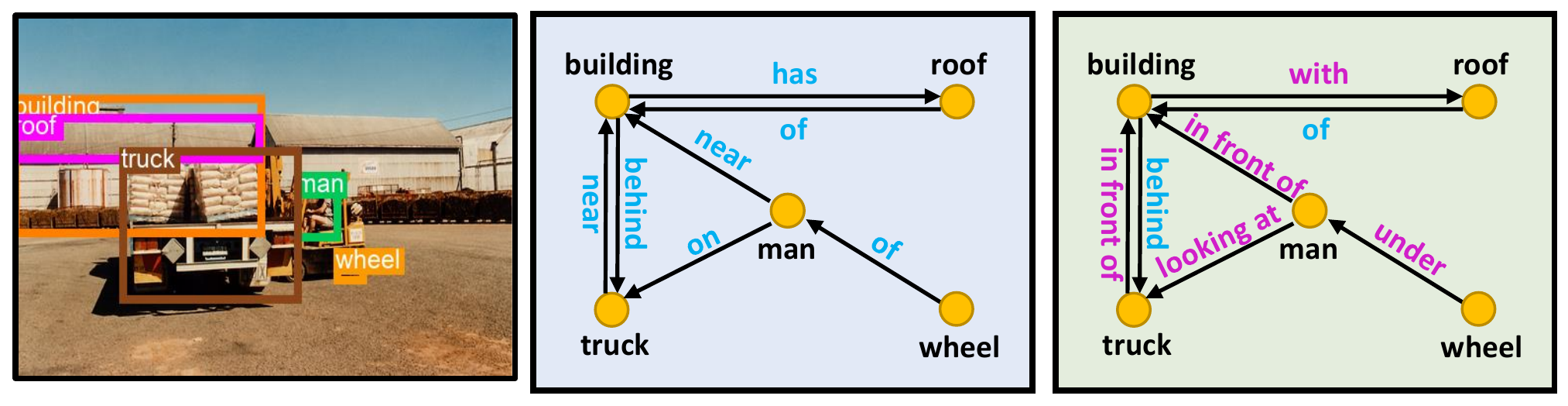}}
	\subfloat[\label{fig:7b}]{
		\includegraphics[width=0.48\textwidth]{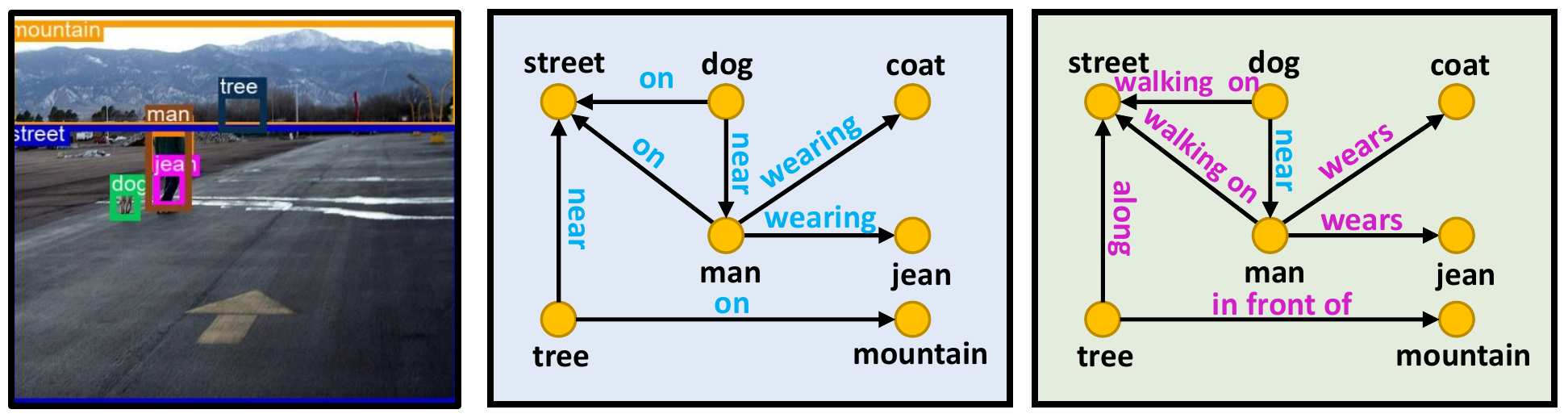}}
	\\
	\subfloat[\label{fig:7c}]{
		\includegraphics[width=0.48\textwidth]{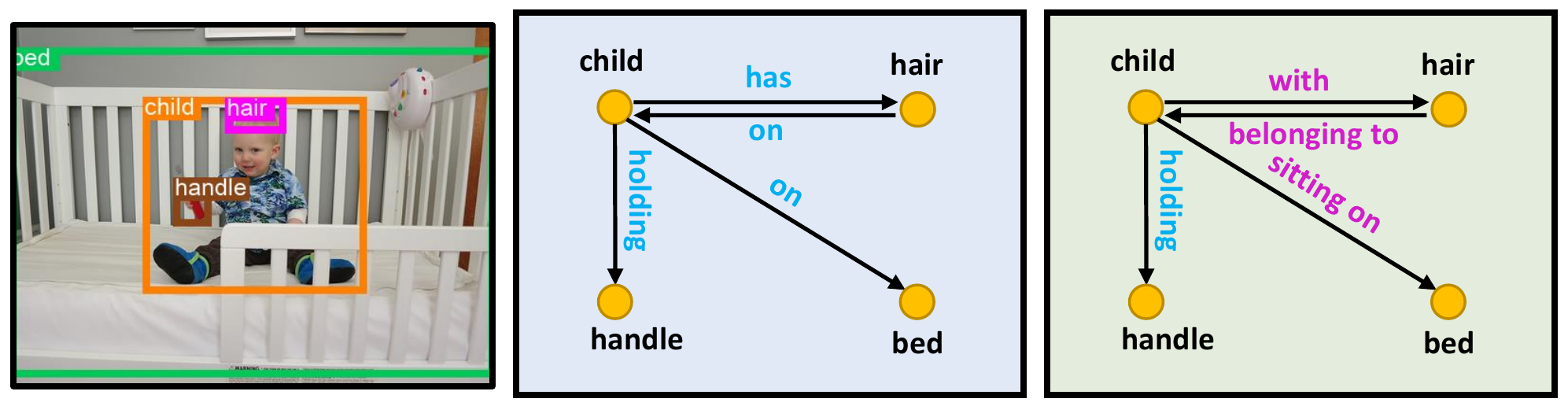}}
	\subfloat[\label{fig:7d}]{
		\includegraphics[width=0.48\textwidth]{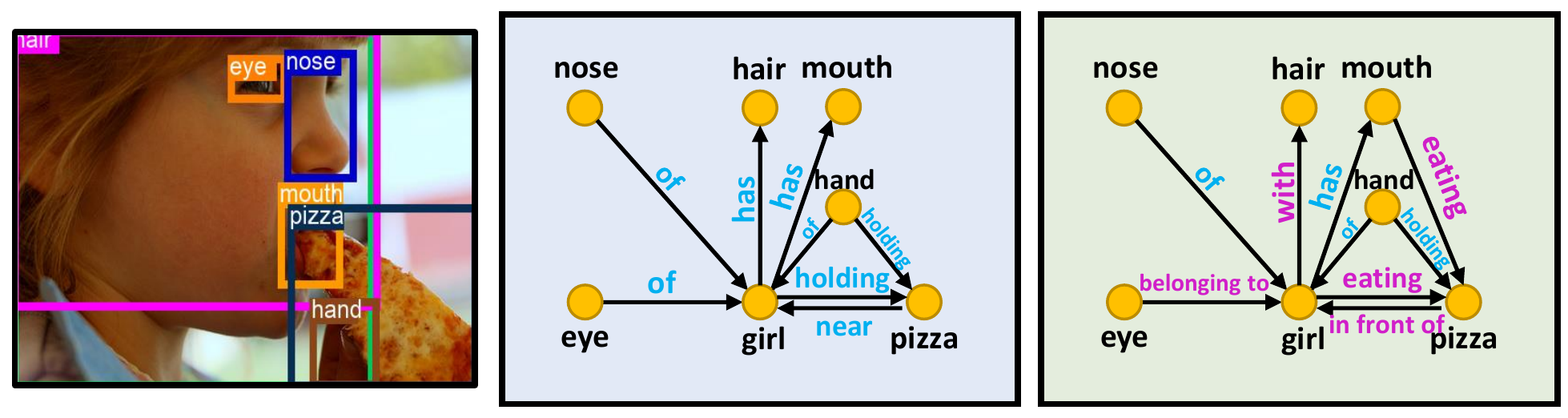}}
	\caption{Comparison between our TISGG and the baseline model VCTree \cite{tang2019learning} on four images. The blue background scene graphs are from VCTree, and the green background scene graphs are from our TISGG. The purple predicates are the corrections of our model to the baseline results.}
		\label{fig:qualitive}
\end{figure*}

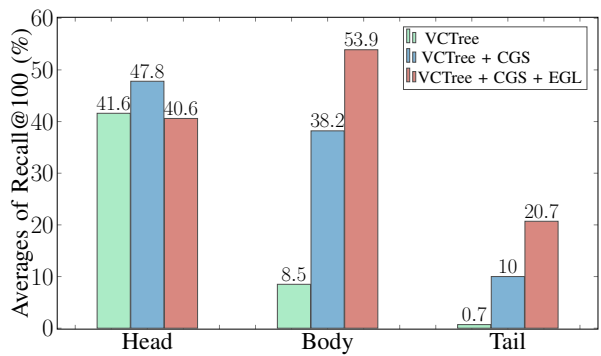
\begin{figure}[t]
\centering
\definecolor{t}{RGB}{171,235,198}
\definecolor{gs}{RGB}{127,179,213}
\definecolor{cr}{RGB}{217,136,128}
\definecolor{black}{rgb}{0.3,0.3,0.3}
\begin{tikzpicture}[scale=0.38]
\begin{axis}[
height = 12.5cm,
width = 20.5cm,
ymax=60,
ymin=0,
x tick label style = {font=\Huge},
y tick label style = {font=\Huge},
ylabel style = {font=\Huge,yshift=0pt},
ylabel=Averages of Recall@100 (\%),
enlarge x limits =0.25,
enlarge y limits = 0.003,
legend style={area legend,at={(0.81,0.97),font=\LARGE},anchor=north, legend columns=1},
ybar=0.5pt,
bar width=33pt,
point meta=y,
tick align=inside, 
xticklabels={,,,Head,,,,,Body,,,,,Tail},
nodes near coords,
nodes near coords style={font=\huge},
nodes near coords align = {vertical}
]
\addplot[draw=black,fill=t]
coordinates {(0,41.6) (1,8.5) (2,0.7)};
\addplot[draw=black,fill=gs]
coordinates {(0,47.8) (1,38.2) (2,10.0)};
\addplot[draw=black,fill=cr]
coordinates {(0,40.6) (1,53.9) (2,20.7)};
\legend{VCTree \ \ \ \ \ \ \ \ \ \ \ \ \ \ \ \   , VCTree + CGS\ \ \ \ \ \ \ \ \  , VCTree + CGS + EGL }
\end{axis}
\end{tikzpicture}
\caption{The figure shows the averages of R@100 of the head, body, and tail classes of predicates using different balancing strategies. The three classes contain 16, 17, and 17 categories of predicates, respectively.}
\vspace{-1em}

\label{fig:CR}
\end{figure}

\subsection{Ablation Study}

For a clear and concise presentation of our model performance, we focus on the results of our model in the PredCls subtask. We have chosen VCTree-based settings in this section and done independent experiments on each module to verify their effectiveness. As shown in Table \ref{tab:ablation}, each module improves zR@K and mR@K when implemented individually, indicating that all modules can help constructing novel unseen triples, and model performance is further improved when the modules work together. 

\noindent
\textbf{Independent Component.}
From Table \ref{tab:ablation}, we can see that the FKR module has significantly improved the generalization ability of VCTree, more than twice at zR@20/50, so independent learning and factual knowledge can significantly improve the generalization ability of the model. Also, the feature learning strategies such as joint feature learning (JFL) and guided sampling (CGS) can help constructing novel triples. IR, considers the information content of the predicates, can reduce the bias of the head predicates to some extent, but not as good as other components. 
At the same time, we found that JFL can help predicting many challenging predicates. For example, the recalls of \emph{made of}, \emph{growing on}, and \emph{painted on} were always 0 in the existing models, they are now increased to 0.0312, 0.2160, and 0.0591 in JFL prediction. It can be seen that our model is less affected by the biased datasets after text and visual features alignment, and the model can correctly predict many difficult tail predicates.
CGS can alleviate the long-tailed effect in the dataset, so the bias in the triple constructions process is reduced. As shown in Table \ref{tab:ablation}, CGS itself can improve zR@K. In terms of the long-tailed problem, CGS constructs a relatively balanced target domain with a reasonable predicate distribution, so the long-tailed problem is effectively mitigated, and the mR@K results of the model surge with the help of CGS. Notably, it can be seen in Figure \ref{fig:CR} that the recalls of head predicate in the blue bars are higher than the green ones. CGS not only improves the recall of body and tail predicates but also protects some of the challenging head predicates and improve the corresponding recalls, so the first blue bar is even higher than the first green bar.
In addition, as can be seen from Table \ref{tab:ablation}, IR and CGS significantly affect the improvement of mR@K, which indicates that these two methods can effectively alleviate the long-tailed problem, and balanced learning positively impacts the model's generalization ability. Other methods can also overcome the long-tailed problem to some extent after increasing the independence of the model, so FKR and JFL can also improve mR@K.

\noindent
\textbf{Cooperate Components.}
In order to eliminate the ambiguity we mentioned earlier, we add the joint learning network to FKR (FKR + JFL). We can see that the model can construct more unseen triples, and many errors caused by ambiguity in the original output were solved. In addition, the results of mR@K in the table show that, when the factual knowledge and feature learning helps the model to predict independently, the long-tailed problem brought by the original dataset can be alleviated.
The results in the table show that JFL has a positive impact when working with the three different modules, suggesting that joint feature learning is very effective and can be adapted to various learning methods. CGS also performs well when working with other modules, suggesting that other modules can perform better when provided with a reasonable target domain.
From Table \ref{tab:ablation}, we can see that when IR cooperate with FKR, the results decrease compared with other settings. We believe this is because both IR and FKR make refinements to predicate classification, but they sometimes refine in different aspects. IR favors predicates with high information content, and FKR favors predicates with stronger object relations, so they sometimes conflict to each other. However, it is undeniable that equipping IR only in SGG can also improve the model's generalization ability. On the other hand, IR is a helpful strategy to alleviate the long-tailed problem, and when it works with CGS, the long-tailed problem is alleviated to a great extent. As shown in Figure \ref{fig:CR}, the recalls of the body and tail predicates are greatly improved, and the prediction results of the model are more balanced.

\subsection{Qualitative Results}
Here we show some qualitative results obtained by the proposed TISGG model. Figure \ref{fig:qualitive} shows that our method tends to generate more reasonable predicates than VCTree, thus making the model more specific and informative. For example, in Figure \ref{fig:7a}, \ref{fig:7c}, and \ref{fig:7d}, our model predict \emph{with} instead of \emph{has} (between \emph{building} and \emph{roof}, \emph{child} and \emph{hair}, and \emph{girl} and \emph{hair}). Indicating that, on the one hand, the semantic information brought by our feature learning makes the text more fluent. On the other hand, our CGS successfully protects the complex predicates we mentioned before. In Figure \ref{fig:7a}, we predict \emph{in front of} (between \emph{man} and \emph{building}) instead of \emph{near}, and \emph{under} (between \emph{wheel} and \emph{man}) instead of \emph{of}. These results show that our model has mitigated the long-tailed problem and the predictions are more specific (Similarly in Figure \ref{fig:7d}). Besides, the utilization of factual knowledge allows the model to analyze scenarios more contextually rather than simply predicting the head predicates. For instance, in Figure \ref{fig:7b}, VCTree outputs the triple \textit{\textless man, on, street\textgreater}, while TISGG can correctly predict the relation between \emph{man} and \emph{street} as \emph{walking on}. It is worth noting that text-image alignment feature learning can also be beneficial for error correction. In Figure \ref{fig:7b}, the baseline model predict \textless tree, on, mountain\textgreater, but we can correctly construct \textless tree, in front of, mountain\textgreater according to the visual information. These examples show that our model gains certain linguistic expression abilities, making the predicted triples smoother. It can be seen from the figure that our model produces more general and sophisticated scene graphs with rational and comprehensive triples.

\section{Conclusion}
We propose a Text-Image alignment Scene Graph Generation (TISGG) that utilizes text and image joint features and factual knowledge of the individual element of the triples. We align the image and the corresponding text features,  reduce the ambiguity caused by polysemies, and jointly learn the visual and semantic information. We introduce factual knowledge of the object and predicate categories, so that the model can be independent of datasets to some extent. Meanwhile, we propose a tailor-made sampling strategy based on the characters of the predicates to reduce the impact of biased data on generalization ability. We evaluate our TISGG on different datasets. From the experimental results, we can see that TISGG performs well on multiple settings. The results under the zero-shot setting have been greatly improved, our model achieves state-of-the-art performance, and the model's generalization ability is substantially improved.

\bibliographystyle{IEEEtran}
\bibliography{tisgg}

\end{document}